\pdfoutput=1
\documentclass[11pt]{article}

\usepackage{acl}

\usepackage{times}
\usepackage{latexsym}

\usepackage[T1]{fontenc}
\usepackage[utf8]{inputenc}

\usepackage{microtype}

\usepackage{inconsolata}

\usepackage{graphicx}
\usepackage{booktabs}
\usepackage{multirow}
\usepackage{amssymb}
\usepackage{amsmath}
\usepackage{bbm}
\usepackage{algorithm}
\usepackage{algorithmic}
\usepackage{xcolor}
\usepackage{verbatim}
\definecolor{light-gray}{gray}{0.95}
\newcommand{\code}[1]{\colorbox{light-gray}{\texttt{#1}}}
\usepackage{todonotes}
\usepackage{subcaption}
\usepackage{listings}

\title{Conformal Intent Classification and Clarification for Fast and Accurate Intent Recognition}

\author{Floris den Hengst \\
  Vrije Universiteit Amsterdam \\
  \texttt{f.den.hengst@vu.nl} \\\And
  Ralf Wolter \\
  ING Group NV \\
  \texttt{Ralf.Wolter@ing.com} 
  \\\AND
  Patrick Altmeyer \\
  TU Delft \\
  \texttt{P.Altmeyer@tudelft.nl} 
  \\\And  
  Arda Kaygan \\
  ING Group NV \\
  \texttt{Arda.Kaygan@ing.com}
  }

\begin{document}
\maketitle
\begin{abstract}
We present Conformal Intent Classification and Clarification (CICC), a framework for fast and accurate intent classification for task-oriented dialogue systems. The framework turns heuristic uncertainty scores of any intent classifier into a clarification question that is guaranteed to contain the true intent at a pre-defined confidence level.
%The user is asked to select their true intent from the set with a clarification question. %We propose to formulate this question with a generative language model.
By disambiguating between a small number of likely intents, the user query can be resolved quickly and accurately. Additionally, we propose to augment the framework for out-of-scope detection.
%The clarification question disambiguates between a small number of likely intents so that the user query can be resolved quickly and accurately.
In a comparative evaluation using seven intent recognition datasets we find that CICC generates small clarification questions and is capable of out-of-scope detection.
%CICC uses a modestly-sized calibration set, is agnostic to the intent classification model and does not require retraining.
%With CICC, the challenges of determining when to ask a clarification question and what to include in it, is effectively addressed by selecting from an achievable trade-off between coverage of the true intent and the number of suggested intents in the question.
CICC can help practitioners and researchers substantially in improving the user experience of dialogue agents with specific clarification questions.
\end{abstract}

% NOTE: page limit: 6 pages
\section{Introduction}
Intent classification (IC) is a crucial step in the selection of actions and responses in task-oriented dialogue systems. To offer the best possible experience with such systems, IC should accurately map user inputs to a predefined set of intents. A widely known challenge of language in general, and IC specifically, is that user utterances may be incomplete, erroneous, and contain linguistic ambiguities.

Although IC is inherently challenging, a key strength of the conversational setting is that disambiguation or \emph{clarification} questions (CQs) can be posed~\citep{purver2003means,alfieri2022intent}. Posing the right CQ at the right time results in a faster resolution of the user query, a more natural conversation, and higher user satisfaction~\citep{van2020collecting,keyvan2022approach,siro2022understanding}. CQs have been considered in the context of information retrieval~\cite{zamani2020generating} but have received little attention in the context of task-oriented dialogue. 

Deciding when to ask a CQ and how to pose it are challenging tasks~\citep{devault2007managing,keyvan2022approach}. First, it is not clear when the system can safely proceed under the assumption that the true intent was correctly identified. Second, it is not clear when the model is too uncertain to formulate a CQ~\citep{cavalin2020improving}. Finally, it is unclear what the exact information content of the clarification question should be.

% It appears prudent and practical to incorporate the most likely intents based on systems' own confidence in its classification. How to do so is a formidable challenge in itself, which demands an approach rooted in a robust theoretical framework that not only showcases empirical effectiveness but also offers easily configurable hyperparameters tailored to the specific needs of the use case at hand. This paper aims to propose an approach that meets these criteria and opens the discussion on the broader topic of CQ generation.

We present Conformal Intent Classification and Clarification (CICC), a framework for deciding when to ask a CQ, what its information content should be, and how to formulate it. The framework uses conformal prediction to turn a models' predictive uncertainty into
%rigorous nonconformity scores for individual inputs. Conformal intent classifiers return
prediction sets that contain the true intent at a predefined confidence level~\citep{shafer2008tutorial,angelopoulos2023conformal}. The approach is agnostic to the intent classifier, does not require re-training of this model, guarantees that the true intent is in the CQ, allows for rejecting the input as too ambiguous if the model is too uncertain, has interpretable hyperparameters, generates clarification questions that are small and is amenable to the problem of detecting out-of-scope inputs.
%More ambiguous user utterances are associated with higher uncertainty, which generally leads to larger prediction sets. 

% In the proposed framework, prediction sets trigger clarification questions that can be formulated by e.g. a generative language model (LM). If the set is too large, the input is considered to be too ambiguous for the model and a request to rephrase the question is returned.

In a comparative evaluations with seven data sets and three IC models,
%We evaluate on a data set and model from the banking domain \citep{alfieri2022intent}, on the curated BANKING77 data set by \citet{casanueva2020efficient} with a commercial model called DialogFlowCX and on the BANKING77 data set with a BERT model fine-tuned for IC \cite{kenton2019bert}.
we find that CICC outperforms heuristic approaches to predictive uncertainty quantification in all cases. The benefits of CICC are most prominent for ambiguous inputs, which arise naturally in real-world dialogue settings~\citep{zamani2020generating,larson-etal-2019-evaluation}.

\section{Related Work}
\label{sec:related_work}
We discuss related work on ambiguity and uncertainty detection within IC and CP with language models.

\paragraph{Clarification Questions}
Various works acknowledge the problem of handling uncertainty in intent classification and to address it with CQs. \citet{dhole2020resolving} proposes a rule-based approach for asking discriminative CQs. The approach is limited to CQs with two intents, lacks a theoretical foundation, and provides no intuitive way of balancing coverage with CQ size. \citet{keyvan2022approach} survey ambiguous queries in the context of conversational search and list sources of ambiguity. They mention that clarification questions should be short, specific, and based on system uncertainty. We propose a principled approach to asking short and specific questions based on uncertainty of any underlying intent classifier for the purposes of task-oriented dialogue.

\citet{alfieri2022intent} propose an approach for asking a CQ containing a fixed top-$k$ most likely intents with intent-specific uncertainty thresholds. This approach does not come with any theoretical guarantees and its hyperparameters need to be tuned on an additional data set whereas our approach comes with guarantees on coverage of the true intent and with intuitively interpretable hyperparameters that can be tuned on the same calibration set. We do not compare directly to this method but include top-$k$ selection in our benchmark.

CQs have been studied in other domains, including information retrieval~\cite{zamani2020generating}, product description improvement~\cite{zhang2021diverse}, and open question-answering~\cite{kuhn2022clam}. In contrast to the task-specific domain investigated in this work, these domains leave more room for asking generic questions for clarification and do not easily allow for incorporating model uncertainty. Furthermore, the proposed methods require ad hoc tuning of scores based on heuristic metrics of model uncertainty, and do not provide ways to directly balance model uncertainty with CQ size.

\paragraph{Uncertainty and out-of-scope detection}
The out-of-scope detection task introduced by~\citet{larson-etal-2019-evaluation} is a different task from the task of handling model uncertainty and ambiguous inputs~\citep{cavalin2020improving,yilmaz2020kloos,zhan2021out,zhou2021contrastive}. However, predictive uncertainty is often used in addressing the out-of-scope detection task. Although the tasks of handling ambiguous input and detecting out-of-scope input are different, we briefly discuss approaches that leverage model uncertainty for out-of-scope detection here.

Various out-of-scope detection approaches train an intent classifier and tune a decision boundary based on a measure of the classifier's confidence~\cite{shu2017doc,lin2019deep,yan2020unknown,hendrycks2020pretrained}. Samples for which the predictive uncertainty of the model lies on one side of the boundary are classified as out-of-scope. These approaches use the models' heuristic uncertainty to decide whether an input is out-of-sample whereas we first turn the models' heuristic uncertainty into a prediction with statistical guarantees and then use this prediction to decide when and how to formulate a clarification question. We additionally propose an adaptation of the CICC framework for out-of-scope detection.

\paragraph{Conformal Prediction on NLP tasks}
Conformal Prediction has been used in several NLP tasks, including sentiment classification by \citet{maltoudoglou2020bert}, named-entity recognition by \citet{fisch2022conformal} and paraphrase detection by~\citet{giovannotti2021transformer}. However, the application to intent classification, task-oriented dialogue and the combination with CQs presented here is novel to our knowledge.

\section{Methodology}
\begin{figure*}[tbph]
    \centering
    \includegraphics[width=.8\textwidth]{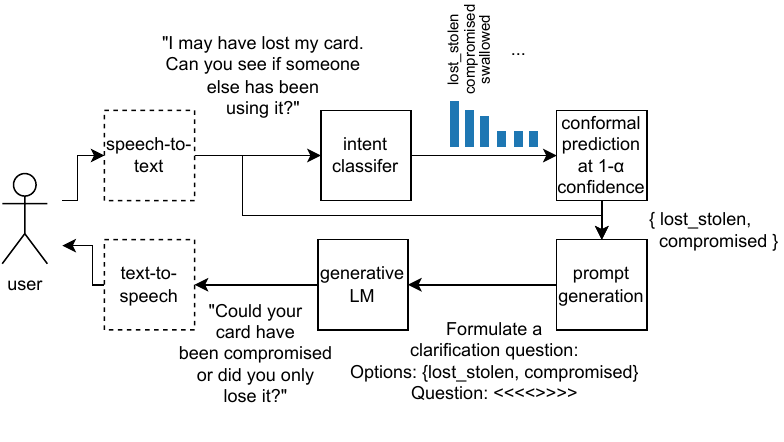}
    \vspace{-.8cm}
    \caption{The conformal intent classification and clarification interaction loop.}
    \label{fig:flowchart}
\end{figure*}

We address the problem of asking CQs in task oriented dialogue systems in the following way.
We take a user utterance and a pre-trained intent classifier, and then return an appropriate response based on the predictive uncertainty of the model. Algorithm~\ref{alg:cicc} lists these steps, and an example input is presented in Figure~\ref{fig:flowchart}. In this section we describe and detail the components of CICC. We start by providing a background on conformal prediction.

\subsection{Conformal Prediction}
\label{sec:cp}
Conformal Prediction is a framework for creating statistically rigorous prediction sets from a heuristic measure of predictive uncertainty of a classifier \citep{shafer2008tutorial,angelopoulos2023conformal}. We here focus on split conformal prediction as it does not require any retraining of the underlying model, and refer to it simply as conformal prediction from here on out.

For a classification task with classes $\mathcal{Y}: \{1,\dots,K\}$, a test input $X_{t}\in\mathcal{X}$ with label $Y_{t} \in \mathcal{Y}$, and a user-defined error level $\alpha \in \left[0,1\right)$, CP returns a set $\mathcal{C}(X_t)\subseteq \mathcal{Y}$ for which the following holds~\cite{vovk1999machine} even when using a finite amount of samples:
\begin{equation}
\label{eq:cp_guarantee}
\mathbb{P}\left(Y_t \in \mathcal{C}(X_t)\right) \geq 1 - \alpha
\end{equation}
If e.g. $\alpha=0.01$ the set $\mathcal{C}(X_t)$ is therefore \emph{guaranteed} to contain the true $Y_t$ in 99\% of test inputs.

Conformal prediction uses a heuristic measure of uncertainty of a pretrained model and a modestly sized calibration set to generate prediction sets. Formally, we assume a held-out calibration set $D: \{(X_i, Y_i)\}$ of size $n$, a pre-trained classifier $\hat{f}: \mathcal{X} \to \mathbb{R}^K$, and a nonconformity function $s: \mathcal{X}\times\mathcal{Y}\to\mathbb{R}$ that returns heuristic uncertainty scores where larger values express higher uncertainty. An example of a nonconformity function for a neural network classifier is one minus the softmax outputs of the true class:
\begin{equation}
    \label{eq:marginal_cp}
    s(X_i) := 1-\hat{f}(X_i)_{Y_i}.
\end{equation}
This score is high when the softmax score of the true class is low, i.e., when the model is badly wrong.

The nonconformity function $s$ is evaluated on $D$ to generate a set of nonconformity scores $\mathcal{S} = \{s(X_i, Y_i)\}$. Next, the quantile $\hat{q}$ of the empirical distribution of $\mathcal{S}$ is determined so that the desired coverage ratio $(1-\alpha)$ is achieved. This can be done by choosing $\hat{q}=\lceil(n+1)(1-\alpha)\rceil/n$\footnote{this is essentially the $\hat{q}$ quantile with a minor adjustment} where $\lceil\cdot\rceil$ denotes the ceiling function. Then, for a given test input $X_t$, all classes $y \in \mathcal{Y}$ with high enough confidence are included in a prediction set $\mathcal{C}(X_t)$ :
\begin{equation}
    \label{eq:prediction_set}
    \mathcal{C}(X_t) := \{y: s(X_t, y) \leq \hat{q}\}.
\end{equation}
This simple procedure guarantees that (\ref{eq:cp_guarantee}) holds i.e. that the true $Y_t$ is in the set at the specified confidence $1-\alpha$. Note the lack of retraining or ensembling of classifiers, that the procedure requires little compute and that $D$ can be relatively small as long as it contains a fair number of examples for all classes and is exchangeable\footnote{distributed identically but not necessarily independently} with the test data~\cite{papadopoulos2002inductive}.

% CP calibrates these heuristic scores into statistically rigorous prediction sets on a held-out calibration set $D$ that is identically distributed to the test set. It does so by evaluating a pre-trained model $\hat{f}$ on the calibration set $D$ This set can be relatively small but should contain a fair number of examples for all classes.

% The CP frameowork

% [LIFTED FROM https://www.paltmeyer.com/blog/posts/conformal-prediction/]

% 1. Partition the training into a proper training set and a separate calibration set: $\mathcal{D}_n=\mathcal{D}^{\text{train}} \cup \mathcal{D}^{\text{cali}}$.
% 2. Train the machine learning model on the proper training set: $\hat\mu_{i \in \mathcal{D}^{\text{train}}}(X_i,Y_i)$.
% 3. Compute nonconformity scores, $\mathcal{S}$, using the calibration data $\mathcal{D}^{\text{cali}}$ and the fitted model $\hat\mu_{i \in \mathcal{D}^{\text{train}}}$. 
% 4. For a user-specified desired coverage ratio $(1-\alpha)$ compute the corresponding quantile, $\hat{q}$, of the empirical distribution of nonconformity scores, $\mathcal{S}$.
% 5. For the given quantile and test sample $X_{\text{test}}$, form the corresponding conformal prediction set: 

% $$
% C(X_{\text{test}})=\{y:s(X_{\text{test}},y) \le \hat{q}\}
% $$ 

% \subsection{Evaluation Measures}
There are various implementations of conformal prediction with different nonconformity functions and performance characteristics.
The most simple approach is known as \emph{marginal} conformal prediction and it uses the nonconformity function in (\ref{eq:marginal_cp}). Marginal conformal prediction owes its names from adhering to the guarantee (\ref{eq:cp_guarantee}) marginalized over $\mathcal{X}$ and $\mathcal{Y}$, i.e. it satisfies the coverage requirement~(\ref{eq:cp_guarantee}) on average, rather than e.g. for a particular input $X_t$. 
Marginal CP can be implemented following the steps described previously: (i) compute nonconformity scores $S$ using (\ref{eq:marginal_cp}), (ii) obtain $\hat{q}$ as described previously, and (iii) construct a prediction set using (\ref{eq:prediction_set}) at test time. A benefit of this approach is that it generates prediction sets with the smallest possible prediction set size on average. A limitation is that its prediction set sizes may not reflect hardness of the input~\cite{sadinle2019least}.

Alternatively, one can ensure conditional adherence to (\ref{eq:cp_guarantee}) with so-called conditional or adaptive conformal predictors. A benefit of conditional approaches is that higher model uncertainty results in larger prediction sets. However, a downside is that these sets are expected to be larger on average than those obtained with a marginal approach.\citet{romano2020classification} introduce a conditional CP approach that consists of broadly the same steps as marginal CP but with a different nonconformity function $s$ and a different prediction set construction. First we define a permutation $\pi(X)$ of $\{1\dots K\}$ that sorts $\hat{f}(X)$ in descending order. Conditional CP can defined as: (i) sum all predictor outputs $\hat{f}(X_i)_k$ for all $\{k \in K|\hat{f}(X_i)_k \geq \hat{f}(X_i)_{Y_i}\}$, (ii) obtain $\hat{q}$ as before, and (iii) include all for a test input $X_t$:
\begin{equation}
\label{eq:conditional_prediction_sets}
\mathcal{C}(X_t) := \{\pi_1(X_t),\dots,\pi_k(x)\},
\end{equation}
where
\begin{equation}
    \label{eq:k_conditional_ps}
    k=\text{sup}\left\{ k': \sum_{j=1}^{k'} \hat{f}(X_t)_{\pi_j(X_t)} < \hat{q} \right\} + 1.
\end{equation}

\citet{angelopoulos2020uncertainty} introduce an approach with a term to regularize the prediction set size: their approach is therefore known as Regularized Adaptive Prediction Sets (RAPS). It effectively adds an increasing penalty to the ranked model outputs in the first step of conditional CP in order to promote smaller prediction sets where possible. Since the second and third step are similar to conditional CP, its prediction sets still adhere to the coverage guarantee~(\ref{eq:cp_guarantee}).

In general, a suitable conformal prediction technique strikes the right balance between three desiderata: (i) adhering to the coverage requirement in (\ref{eq:cp_guarantee}), (ii) producing small prediction sets and (iii) adaptivity. Whereas the former two can be measured easily, metrics for adaptivity require some more care. \citet{angelopoulos2020uncertainty} introduce a general-purpose metric for adaptivity. It is based on the coverage and referred to as the size-stratified classification (SSC) score:
\begin{equation}
\label{eq:ssc}
\text{SSC} = \min_{b \in \{1,\dots,K\}} \frac{1}{|\mathcal{I}_b|} \sum_{i\in\mathcal{I}_b} \mathbbm{1}\left\{ Y_i \in \mathcal{C}\left(X_i\right)\right\}
\end{equation}
for a classification task defined as above and $\mathcal{I}_b \subset \{1,\dots,n\}$ the set of inputs with prediction set size $b$, i.e. $\mathcal{I}_b := \left\{X_i, |\mathcal{C}(X_i)|=b\right\}$.

Within CICC, conformal prediction is applied to a pre-trained intent classifier to create a set of intents that contains the true user intent at a predefined confidence for any user utterance. The sets are then used in making a decision on when to ask a clarification question and how to formulate it. We continue to discuss when and how such questions are asked based on Algorithm~\ref{alg:cicc} in the following section.
\begin{algorithm}[tb]
\caption{CICC algorithm}
\label{alg:cicc}
\begin{flushleft}
\textbf{Input}: utterance $X$, classifier $\hat{f}$, chat/voice-bot $c$, calibration set $D$, generative LM $g$ \\
\textbf{Parameters}: error rate $\alpha$, threshold $th$, ambiguity response $a$         \\
\textbf{Output}: response $R$
\end{flushleft}
\begin{algorithmic}[1] %[1] enables line numbers
\STATE set $\gets$ conformal prediction$\left(\hat{f}(X), D, \alpha\right)$
\IF{$|$set$| == 1$}
\STATE $R \gets c($set.get()$)$.  \hfill \COMMENT{bot response}
\ELSIF{$|$set$| > th$}
\STATE $R \gets a$. \hfill \COMMENT{input too ambiguous}
\ELSE
\STATE $R \gets g($set$, X)$ \hfill\COMMENT{clarification question}
\ENDIF  
\end{algorithmic}
\end{algorithm}

% Description of conformal prediction framework. Focus on split-conformal prediction due to not having to retrain etc.

% Description of prediction set size, marginal coverage and conditional coverage. Description of class-conditional coverage as generic metric for measuring adaptivity.

% Description of specific algorithms: marginal, conditional (cumulated score), RAPS and TOPK.
\subsection{When to Ask a Clarification Question}
For a user utterance $X$, a pre-trained intent classifier $\hat{f}$ and a nonconformity function $s$, we generate a prediction set that covers the true user intent with confidence $1-\alpha$ (Algorithm~\ref{alg:cicc}, ln 1). If the set contains a single intent, the model is confident that the true intent has been detected and the dialogue can be handled as usual (ln 2-3).

If the set contains many intents, that is, more than a user-specified threshold $th \in \mathbb{N}_{>0}$, then there is no reasonable ground for formulating a clarification question. Instead, a generic request to rephrase the question can be asked (ln 4-5), or a hand-over to a human operator could be implemented here. In the remaining case, i.e. if the prediction set is of reasonable size, a CQ is asked (ln 6-7).
%CICC can be used in conjunction with any arbitrary approach to stop the conversation if, e.g., too many repeated questions to rephrase or clarify are posed in succession, by wrapping Algorithm~\ref{alg:cicc} in an algorithm that implements use-case specific stopping conditions.

CICC comes with two parameters to control when a CQ should be asked. Both have clear semantics and can be interpreted intuitively. The first is the threshold $th$ that controls when the input is too ambiguous to ask a CQ (Algorithm~\ref{alg:cicc} ln 4-5). This parameter is set by the chatbot owner on the basis of best practices in, and knowledge of chat- and voicebot interaction patterns. In general, this number should remain small to reduce the cognitive load on users. We advise to set this value no higher than seven~\cite{miller1956magical,plass2010cognitive}.

The second parameter is the error rate $\alpha$. It controls the trade-off between the prediction set size and how certain we want to be that the prediction set covers the true intent. As $\alpha \to 0$, our confidence that the true intent is included in the set grows, but so does the size of the prediction set. Because conformal prediction is not compute intensive, $\alpha$ can be set empirically. Thus, CICC provides a means of selecting between \emph{achievable} trade-offs between prediction set sizes and error rates. We continue to discuss how specific CQs are formulated in CICC.

\subsection{Generating a Clarification Question}
When a CQ is in order (ln 6-7 in Alg.~\ref{alg:cicc}), it needs to be formulated. We propose to generate a CQ based on the original input $X$ and the prediction set, as it is guaranteed to contain the true intent at a typically high level of confidence. Because the alternatives in the CQ are the most likely intents according to the model, and because the number of alternatives in the CQ corresponds to the models' uncertainty, asking a CQ provides a natural way of communicating model uncertainty to the user while quickly determining the true user intent.

CICC makes no assumptions about the approach for generating a CQ. Anything from hardcoded questions, templating, or a generative LM can be used. However, we recognize that the number of possible questions is large: it consists of the powerset of all $n$ intents up to size $th$ excluding sets of size one and zero. Therefore, we opt to use a generative LM in our solution.

We prompt the LM to formulate a clarification question by giving it some examples of clarification questions for a set of example intents to disambiguate between. We additionally provide the original utterance $X$ to enable the formulation of CQ relative to the original utterance. See Appendix~\ref{sec:appendix_implementation_details} for details.

% \subsection{Conformal Prediction Approaches}
% Three desiderata are considered in the evaluation of CP-based solutions.
% First, the prediction sets should \emph{cover} the ground truth up to the user-specified error rate $\alpha$ in expectation. That is, the ground truth should be included in the prediction set in under $1-\alpha$\% of all samples in the test set. Second, the prediction sets should be as small as possible.
% Third, the size of the prediction set should reflect the complexity of the input. Approaches that achieve the latter are referred to as \emph{adaptive}. 

\subsection{Out-of-scope Detection}
Ambiguity is a part of natural language which could lead to model uncertainty. Specific reasons for uncertainty in intent recognition are inputs that are very short and long, imprecise and incomplete inputs, etc. However, a particularly interesting type of uncertainty stems from inputs that represent intent classes that are not known at training time~\cite{zhan2021out}. These inputs are referred to as out-of-scope (OOS) and detecting these inputs can be seen as a binary classification task for which data sets with known OOS samples have been developed.
%OOS is a challenging problem of practical importance in task-oriented dialogue systems.

CICC rejects inputs about which the model is too uncertain (Algorithm~\ref{alg:cicc}, ln 5) and this naturally fits with the OOS detection task as follows: we can view a rejection of an input as a classification of that input as OOS. Therefore, although handling ambiguity in the model gracefully and detection OOS inputs are separate challenges, vanilla CICC implements a form of OOS detection.

Additionally, the CICC framework can be leveraged for OOS detection if OOS samples are known at calibration time. Specifically, we can optimize parameters $\alpha$ and $th$ to maximize predictive performance expressed by some suitable metric such as the F1-score on the calibration set. OOS samples can be obtained from other intent recognition data sets in other domains. This practice is described in detail by e.g. ~\cite{zhan2021out} under the name of open-domain outliers. We refer to versions of CICC which have been optimized for F1-score in this way as CICC-OOS.

\section{Experimental Setup}
This section lists the experiments performed to comparatively evaluate CICC across seven data sets and on three IC models\footnote{
\url{https://github.com/florisdenhengst/cicc}}.
\begin{table}[tp]
\centering
\footnotesize{
\setlength\tabcolsep{2pt}
\begin{tabular}{l|cc}
\toprule
                                                 &    samples    & intents    \\%& $\alpha$\\
\midrule%
ACID \cite{acharya2020using}                    & 22172         & 175               \\%& .02       \\
ATIS \cite{hemphill1990atis}                    & 5871          & 26                \\%& .03        \\
B77 \cite{casanueva2020efficient}               & 13083         & 77                \\%& .03       \\
B77-OOS                                         & 16337         & 78                \\
C150-IS \cite{larson-etal-2019-evaluation}      & 18025         & 150               \\%& .01       \\
C150-OOS  \cite{larson-etal-2019-evaluation}    & 19025         & 151               \\
HWU64 \cite{liu2021benchmarking}                & 25716         & 64                \\%& .05       \\
IND                                             & $\sim$20k     & 61                \\%& .1        \\
MTOD (eng) \cite{schuster2019cross}               & 43323         & 12                \\%& .01       \\
% MTOD-ES \cite{schuster2019cross}                & 8643          & 12                \\
\bottomrule
\end{tabular}
}
\caption{Characteristics of datasets used}
\label{tab:datasets}
\end{table}

\paragraph{Data} We evaluate CICC on six public intent recognition data sets in English and an additional real-life industry data set (IND) from the banking domain in the Dutch language. Table~\ref{tab:datasets} shows the data sets and their main characteristics. All data sets were split into train-calibration-test splits of proportions 0.6-0.2-0.2 with stratified sampling, except for the ATIS data set in which stratified sampling is impossible due to the presence of intents with a single sample. Random sampling was used for this data set instead. We use an in-scope version (C150-IS) of the `unbalanced' data set by~\citet{larson-etal-2019-evaluation} in which all out-of-scope samples have been removed.

For evaluation on out-of-scope (OOS) detection, we use two datasets: a version of C150 with all OOS samples divided over the calibration and test splits, and no OOS samples in the train split (C150-OOS), and a version of B77 with so-called open-domain outliers in which samples from the ATIS dataset make up half of the samples in the calibration and test splits to represent OOS inputs (B77-OOS)~\cite{zhan2021out}.

\paragraph{Models} We employ fine-tuned BERT by \citet{kenton2019bert} for all public data sets and a custom model similar to BERT for the IND data set~\cite{alfieri2022intent}. We base the nonconformity scores on the softmax output in these settings. In order to test performance on a commercial offering, we additionally evaluate using DialogflowCX (DFCX) on the B77 data set.\footnote{\url{https://cloud.google.com/dialogflow/cx/docs}} This commercial offering outputs heuristic certainty scores in the range $ \left[0,100 \right]$ for the top five most certain recognized intents. These outputs were normalized to sum to $1$, all other scores were set to $0$ to determine the nonconformity scores.

\paragraph{Baselines}
In practice CQs can be formulated using heuristics~\cite{alfieri2022intent}. We compare CICC to the following baselines using the models' heuristic uncertainty scores:
\begin{itemize}
\item[B1] select all intents with score $>1-\alpha$, select the top $k=5$ if this selection is empty.
\item[B2] select all intents with a score $>1-\alpha$.
\item[B3] select the top $k=5$ intents.
\end{itemize}

\paragraph{Metrics}
We evaluate the approaches on a set of metrics that together accurately convey the added benefit of asking a confirmation question. We use the \emph{size} of the prediction set $\mathcal{C}(X_i)$ and how often the input is rejected as too ambiguous for the model (Algorithm~\ref{alg:cicc}, ln 5). For a test set of size $n$:
\begin{equation}
    \text{Amb} := \frac{1}{n} \sum_{i=0}^n
        \begin{cases}
            1 & \text{if}~|\mathcal{C}(X_i)| \geq th\\
            0 & \text{otherwise}.
        \end{cases}
\end{equation}

First, we report how often the true intent is detected for the $m \leq n$ inputs that are not rejected (Algorithm~\ref{alg:cicc}, lns 3 and 5). This metric is known as coverage (cov) and can be seen as a generalisation of accuracy for set-valued predictions:
\begin{equation}
    \label{eq:cov}
    \text{Cov} := \frac{1}{m} \sum_{i=0}^m \mathbbm{1}_{\mathcal{C}(X_i)}\left(Y_i\right).
\end{equation}
Second, we report the average size of the clarification questions for accepted inputs (Algorithm~\ref{alg:cicc}, ln 7). This metric can be seen as an analogue to precision for set-valued predictions:
\begin{equation}
    |\text{CQ}| = \frac{1}{m} \sum_{i=0}^m |\mathcal{C}(X_i)|.
\end{equation}
Finally, we report the relative number of times the prediction set is of size one
\begin{equation}
    \text{Single} := \frac{1}{m} \sum_{i=0}^m
        \begin{cases}
            1 & \text{if}~|\mathcal{C}(X_i)| = 1, \\
            0 & \text{otherwise,}
        \end{cases}
\end{equation}
in which case the dialogue can continue as usual (Algorithm~\ref{alg:cicc}, ln 3). We additionally report the SSC as defined above in (\ref{eq:ssc}).

For out-of-scope detection we report the standard metrics F1-score and AUROC.

% \begin{figure*}
%     \centering
%     \begin{subfigure}{.3\textwidth}
%     \centering
%     \includegraphics[width=\columnwidth]{imgs/Industry__multi_crop.png}
%     \end{subfigure}
%     \begin{subfigure}{.3\textwidth}
%     \centering
%     \includegraphics[width=\columnwidth]{imgs/BANKING77_DFCX_multi_crop.png}
%     \end{subfigure}
%     \begin{subfigure}{.3\textwidth}
%     \centering
%     \includegraphics[width=\columnwidth]{imgs/BANKING77_BERT_multi_crop.png}
%     \end{subfigure}
%     \caption{Test set results for varying error rate $\alpha$.}
%     \label{fig:cp_results}
% \end{figure*}

\paragraph{Parameters}
We varied $\alpha$ and found the best settings empirically on the calibration set. We report our key results for the best $\alpha$ and additionally investigate the effect of varying $\alpha$.

We set the threshold $th$ at seven to avoid excessive cognitive load for users for all experiments, except when using DFCX in which case we set $th$ to four~\cite{miller1956magical,plass2010cognitive}. The reason for this is that DFCX currently only outputs non-zero scores for the top five intents. Hence, the set contains all intents that have a non-zero confidence score with this setting.

We include the following conformal prediction approaches and select an approach that produces the best empirical results in terms of coverage and CQ size: marginal, conditional (also known as adaptive) \cite{romano2020classification} and RAPS \cite{angelopoulos2020uncertainty}. Marginal conformal prediction was selected in all experiments, details can be found in Figure~\ref{fig:varying_alpha}.

\section{Results}
\begin{table}[tp]
\centering
\footnotesize{
\setlength\tabcolsep{2pt}
\begin{tabular}{lccr|cccc}
\toprule
Setting & $1-\alpha$ & $th$ &    & Cov$\uparrow$ & Single$\uparrow$ & $|\text{CQ}|\downarrow$ & Amb \\ \midrule
ACID  & .98 & 7 &  CICC & \underline{.98} & .87 & \textbf{3.01} & .03 \\
& & &  B1 & \underline{.98} & \textbf{.88} & 5 & 0 \\
& & &  B2 & .95 & 1 & $-$ & 0 \\
& & &  B3 & \underline{.99} & 0 & 5 & 0 \\
\midrule
ATIS & .99 & 7  &  CICC & \underline{.99} & \textbf{.98} & \textbf{2.54} & 0 \\
& & &  B1 & \underline{.99} & .73 & 5 & 0 \\
& & &  B2 & .98 & 1.00 & - & 0 \\
& & &  B3 & \underline{1.00} & 0 & 5 & 0 \\
\midrule
B77/BERT & .97 & 7 &  CICC & \underline{.98} & .73 & \textbf{2.84} & .04 \\
& & &  B1 & \underline{.97} & \textbf{.84} & 5 & 0 \\
& & &  B2 & .93 & 1 & $-$ & 0 \\
& & &  B3 & \underline{.98} & 0 & 5 & 0 \\
\midrule
B77/DFCX  & .90& 4 &  CICC & \underline{.91} & .66 & \textbf{2.63} & .02 \\
&& &  B1 & \underline{.95} & \textbf{.71} & 5 & .27 \\
&& &  B2 & .90 & .98 & 2.26 & 0 \\
&& &  B3 & \underline{.97} & 0 & 5 & 1 \\
\midrule
C150-ID  & .99 & 7 &  CICC & \underline{.99} & \textbf{.97} & \textbf{2.66} & 0 \\
& & &  B1 & \underline{.99} & .82 & 5 & 0 \\
& & &  B2 & .98 & 1 & $-$ & 0 \\
& & &  B3 & \underline{1} & 0 & 5 & 0 \\
\midrule
HWU64 & .95 & 7 &  CICC & \underline{.95} & \textbf{.82} & \textbf{2.81} & .01 \\
& & &  B1 & \underline{.97} & .70 & 5 & 0 \\
& & &  B2 & .90 & 1 & $-$ & 0 \\
& & &  B3 & \underline{.98} & 0 & 5 & 0 \\
\midrule
IND  & .90 & 7 &  CICC & \underline{.91} & \textbf{.25} & \textbf{3.46} & .11 \\
& & &  B1 & .88 & .42 & 5 & 0 \\
& & &  B2 & .70 & 1 & $-$ & 0 \\
& & &  B3 & \underline{.91} & 0 & 5 & 0 \\
\midrule
MTOD  & .99 & 7  &  CICC & \underline{.99} & \textbf{1} & $-$ & 0 \\
& & &  B1 & \underline{1} & .98 & 5 & 0 \\
& & &  B2 & \underline{.99} & \textbf{1} & $-$ & 0 \\
& & &  B3 & \underline{1} & 0 & 5 & 0 \\
 
\bottomrule
\end{tabular}}
    \caption{Test set results where \underline{underline} indicates meeting coverage requirement. \textbf{Bold} denotes best when meeting this requirement, omitted for last column due to missing ground truth for ambiguous.}
    \label{tab:results}
\end{table}

Table~\ref{tab:results} lists the main results. The first column shows the coverage, i.e. the percentage of test samples in which the ground truth is captured in the prediction set. We see that only CICC and B3 adhere to the requirement of coverage $\geq 1-\alpha$ in all settings. The second column shows the fraction of test samples for which a single intent is detected. We see that CICC outperforms the baselines that meet the coverage requirement in five out of seven data sets.

The third column lists the average size of the CQ. We see that CICC yields the smallest CQs and that the number of inputs that is deemed too ambiguous is relatively small for CICC. The last column denotes the relative number of inputs that is rejected as too ambiguous. CICC rejects a relatively low number of inputs. Upon inspection, many of these inputs could be classified as different intents based on the textual information alone (see Appendix~\ref{sec:appendix_ambiguous_inputs}). For the B77/DFCX setting, we see that B1 predicts a single output frequently, at the cost of rejecting inputs as too ambiguous. This contrasts with CICC, which rejects inputs much less frequently and instead asks a small CQ.

\begin{table}[tp]
\centering
% \footnotesize{
\setlength\tabcolsep{2pt}
\begin{tabular}{llll|cc}
\toprule
Dataset         &  Algorithm    & 1-$\alpha$ & $th$ & F1$\uparrow$    & AUROC$\uparrow$ \\
\midrule
C150-OOS        &   CICC        & .990      & 7     & .07   & .88 \\
                &   CICC-OOS    & .995      & 6     & \textbf{.91}   & \textbf{.97} \\ \midrule
B77-OOS         &   CICC        & .970      & 7     & .76   & .92 \\
                &   CICC-OOS    & .994      & 6     & \textbf{.90}   & \textbf{.97} \\
\bottomrule
\end{tabular}
    \caption{Results for the OOS detection task.}
    \label{tab:ood_results}
\end{table}

\begin{figure*}[tbp]
    \centering
    \begin{subfigure}{.3\textwidth}
    \centering
    \includegraphics[width=\columnwidth]{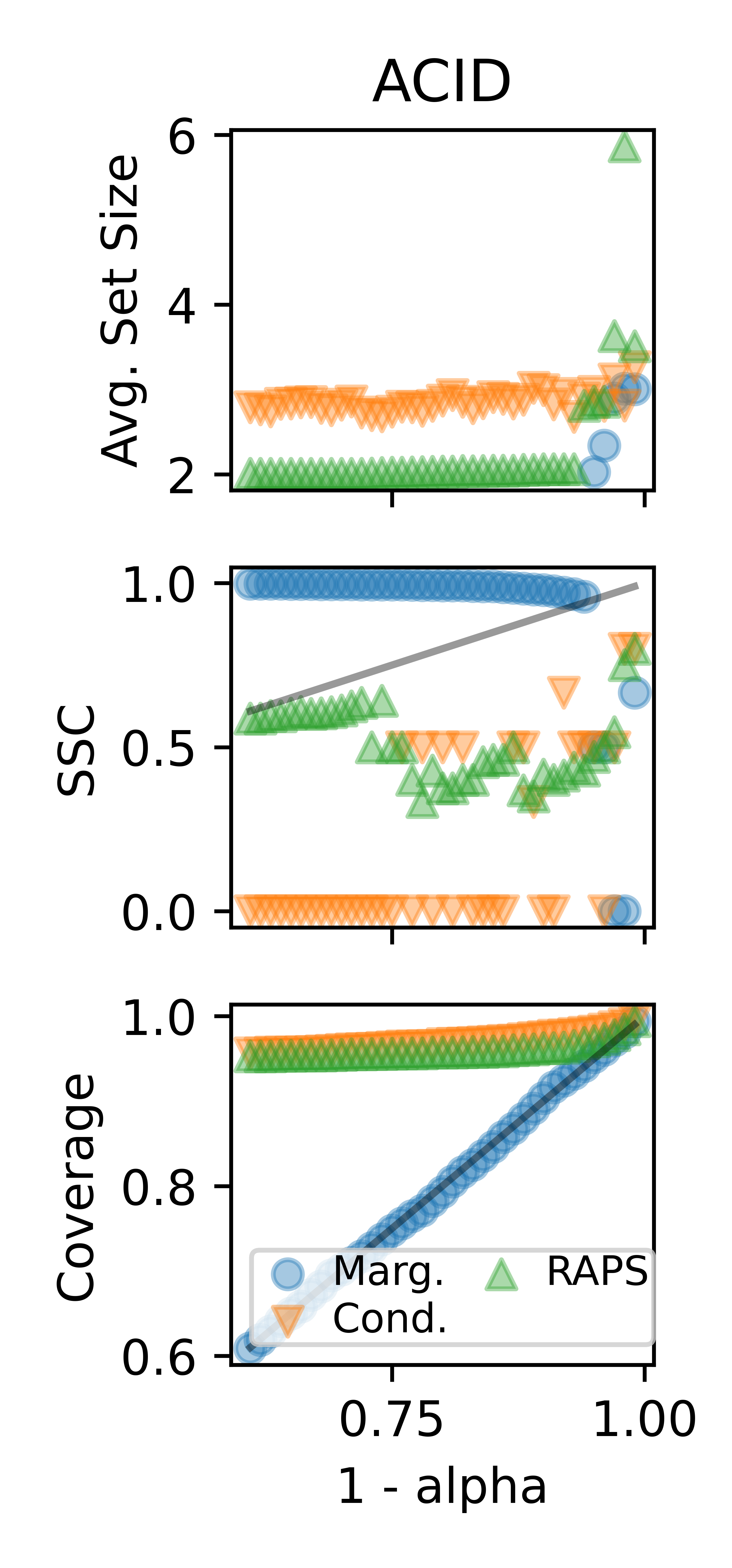}
    \end{subfigure}
    \begin{subfigure}{.3\textwidth}
    \centering
    \includegraphics[width=\columnwidth]{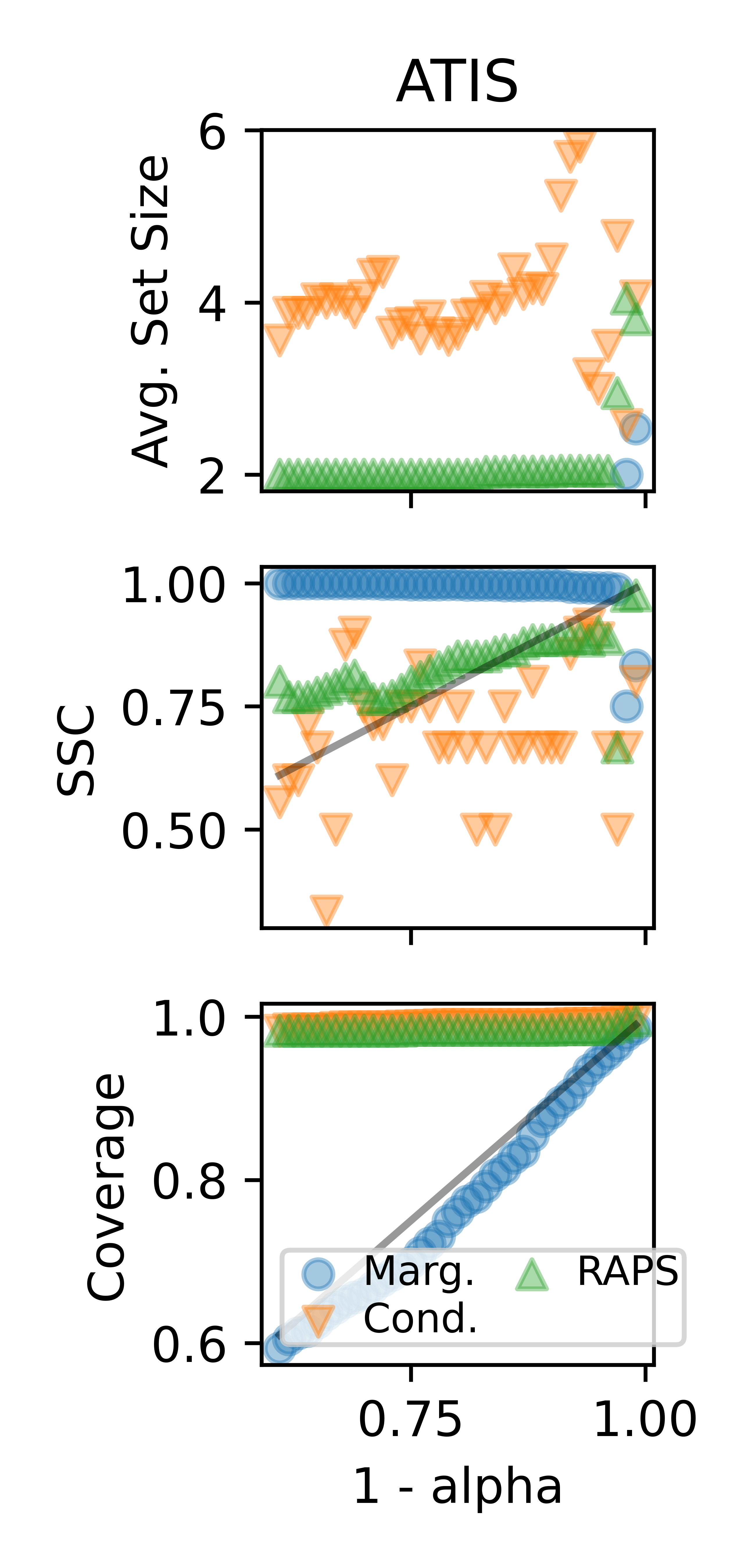}
    \end{subfigure}
    \begin{subfigure}{.3\textwidth}
    \centering
    \includegraphics[width=\columnwidth]{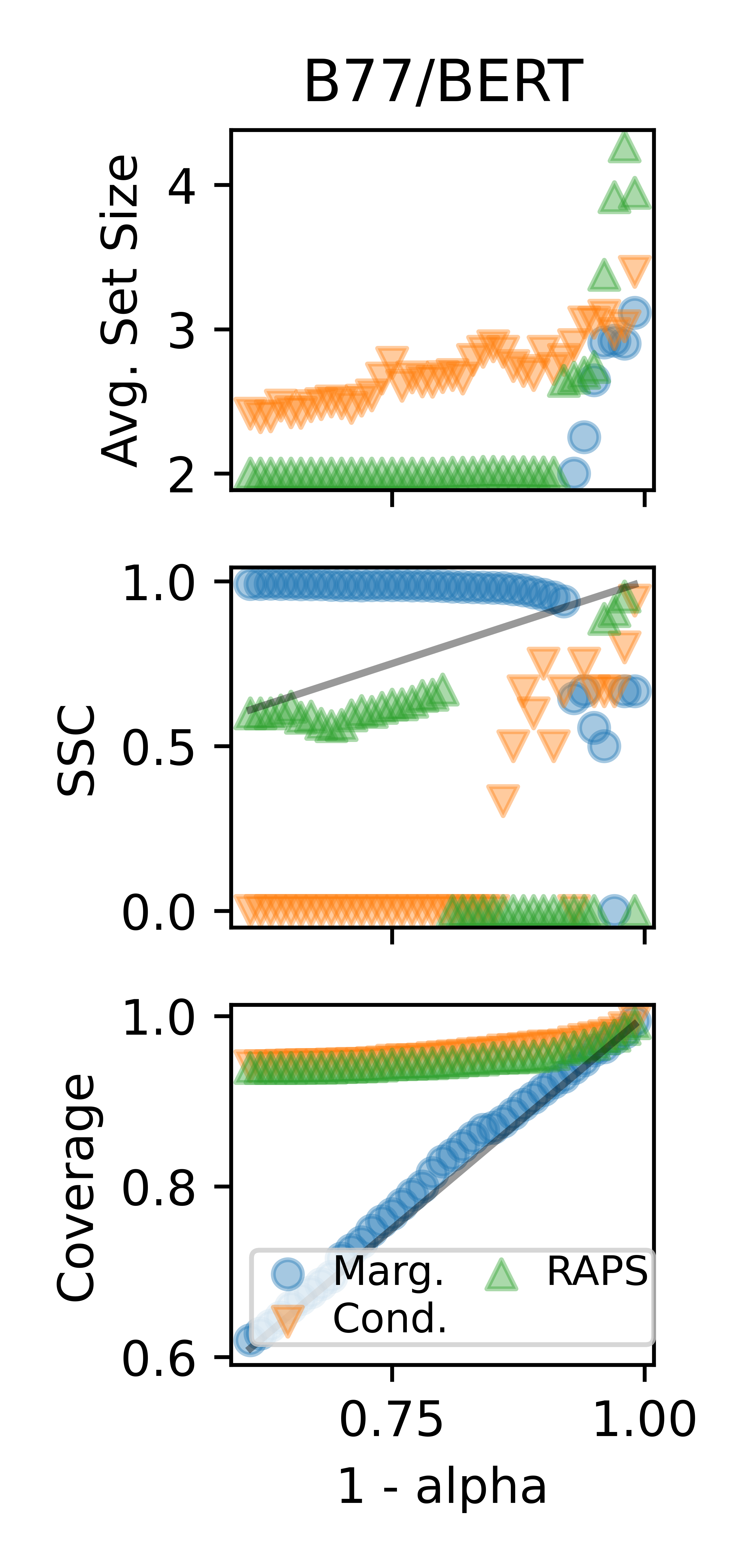}
    \end{subfigure}

    \begin{subfigure}{.3\textwidth}
    \centering
    \includegraphics[width=\columnwidth]{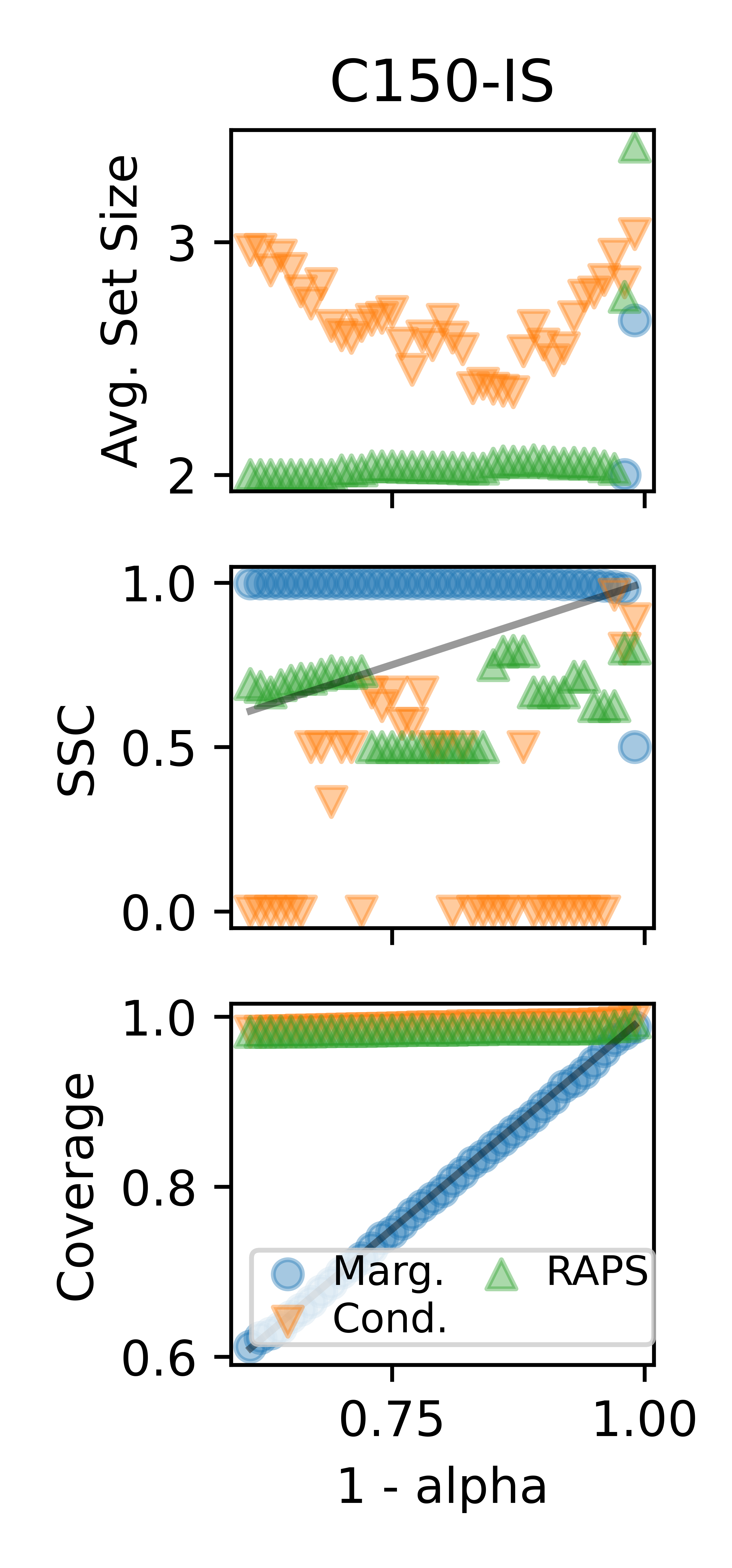}
    \end{subfigure}
    \begin{subfigure}{.3\textwidth}
    \centering
    \includegraphics[width=\columnwidth]{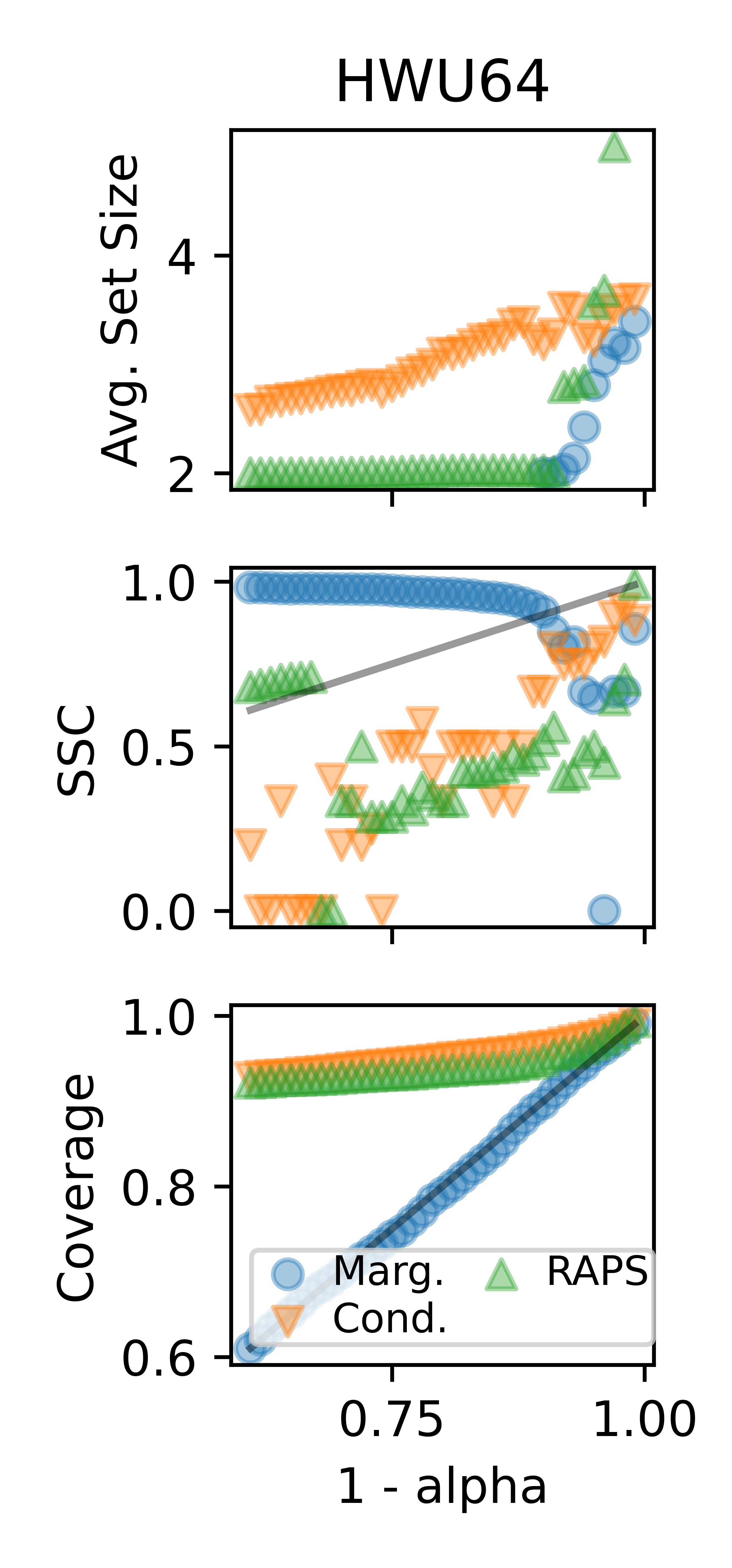}
    \end{subfigure}
    \begin{subfigure}{.3\textwidth}
    \centering
    \includegraphics[width=\columnwidth]{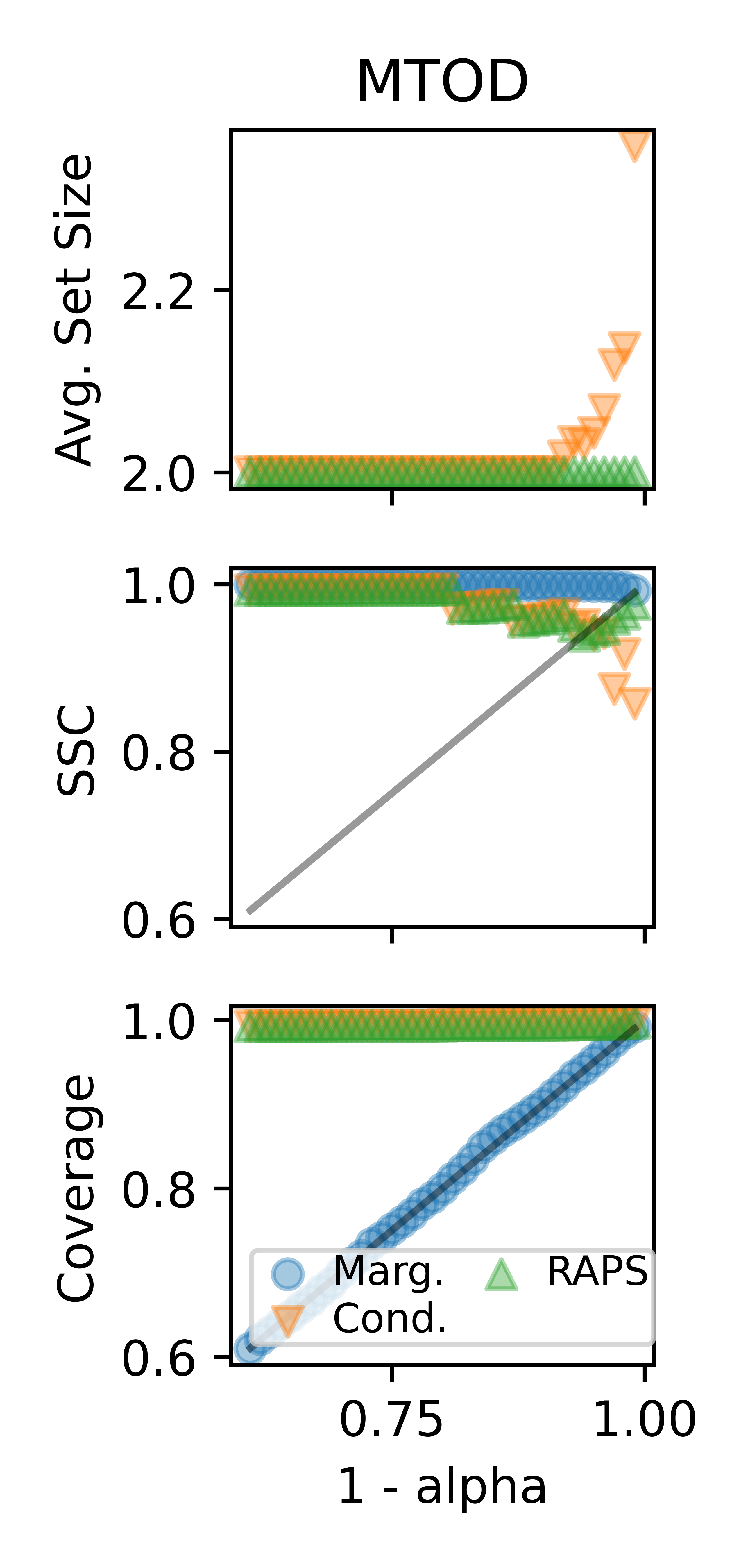}
    \end{subfigure}
    \caption{Test set results for varying error rate $\alpha$.}
    \label{fig:varying_alpha}
\end{figure*}
We continue by looking at the results for OOS detection in Table~\ref{tab:ood_results}. We find that vanilla CICC does not perform well on the OOS detection in comparison to the specialized CICC-OOS variant. The specialized CICC-OOS favours a relatively low $\alpha$ as this simultaneously forces the approach toward large prediction sets for OOS samples and small prediction sets for in-sample inputs. At the same time, using the CICC-OOS settings for parameters $\alpha$ and $th$ in the regular CICC interaction loop would result in relatively many CQs of a relatively large size.

% We continue to look at the ability to detect ambiguous input. Figure~\ref{fig:cumulative_set_sizes} shows the cumulative distribution of the prediction set sizes for the test set. The figures are normalized by the number of intents in the data set to enable a comparison between the data sets. We observe that inputs in the industry setting are more ambiguous and the model used in the public setting is better at recognizing ambiguities than the commercial model within the CICC framework. When investigating the inputs with the largest prediction set sizes, we see that these inputs are indeed open to multiple interpretations, see Appendix~\ref{sec:appendix_ambiguous_inputs}.

Next, we investigate how different conformal prediction approaches perform for varying levels of $\alpha$ in Figure~\ref{fig:varying_alpha}. The top figures show that all conformal prediction approaches enable trading off set size with coverage, a desirable property in practice of intent classification. Looking at the adaptivity (center figures), we see mixed results. A possible explanation for this is in the general-purpose evaluation of adaptivity, which relies on the minimum coverage across classes (see Eq.~\ref{eq:ssc}). The data sets used in our experiments contain a relatively low number of examples for some classes and these rare classes may have an outsized effect on the SSC metric. Looking at the bottom figure for each data set, we see that all conformal prediction approaches lie at or above the x=y diagonal: conformal prediction always adheres to the coverage requirement with the marginal approach yielding the smallest average set sizes. 
%and that the formulation of adaptivity metrics for classification problems with long tails would be highly relevant for the NLP community.

\section{Conclusion}
We have proposed a framework for detecting and addressing uncertainty in intent classification with conformal prediction. The framework empirically determines when to ask a clarification question and how that question should be formulated. The framework uses a moderately sized calibration set and comes with intuitively interpretable parameters.

We have evaluated the framework in eight settings, and have found that the framework strictly outperforms baselines across all metrics in six out of eight cases and performs competitively in the other. The framework additionally handles inputs that are too ambiguous for intent classification naturally. We have additionally proposed and evaluated the usage of CICC for out-of-scope detection and found that it is suitable for this.

We finally believe that the framework opens promising avenues for future work, including the usage of intent groups for better adaptivity, an extension to Bayesian models to address data drift and unsupervised OOS with CICC~\citep{fong2021conformal}, to determine conversation stopping rules based on subsequent questions to rephrase or clarify and to combine it with reinforcement learning for, e.g., personalization~\citep{den2019reinforcement,den2020reinforcement}. We believe that CICC and/or conformal prediction may also prove useful in various other tasks, including entity recognition, detecting label errors~\citep{ying2022label} and to empirically identify similar intents.

% \section*{Ethics Statement}
% Scientific work published at EMNLP 2023 must comply with the \href{https://www.aclweb.org/portal/content/acl-code-ethics}{ACL Ethics Policy}. We encourage all authors to include an explicit ethics statement on the broader impact of the work, or other ethical considerations after the conclusion but before the references. The ethics statement will not count toward the page limit (8 pages for long, 4 pages for short papers).

% \clearpage
\section*{Limitations}
A limitation of the framework is that it relies on a user determining values for the hyperparameters $\alpha$ and $th$. The former balances model certainty with CQ size. Arguably, this trade-off has to be made in any approach and CICC makes this an explicit choice between achievable trade-offs. The threshold $th$ must be set not to reject too many inputs as too ambiguous while avoiding information overload in the user. We advise setting it to no more than seven based on established insights from cognitive science~\cite{miller1956magical}. However, more research on the impact of CQ size on user satisfaction in various context is in order. Another limitation is that the approach does not include a mechanism for stopping the dialogue. We leave the investigation of stopping criteria based on e.g. the number and size of CQs asked during the dialogue for future work. Furthermore, this work did not thoroughly investigate the quality of the CQs produced by the LLM. However, we view the CQ production component as a pluggable component and therefore believe a full-scale evaluation on this to be out-of-scope for this work. Additionally, using CICC for OOS detection requires the presence of OOS labels. While these can be obtained from other data sets using the practice of open-domain outliers~\cite{zhan2021out}, fully unsupervised approaches based on e.g. hierarchical Bayesian modeling or with parameters that yield good performance across data sets as hinted at by~Table~\ref{tab:ood_results}. A final limitation is that we applied conformal prediction to the softmax of outputs of uncalibrated neural network outputs. This makes results consistent across settings (including DFCX), but smaller CQs may be achievable by applying Platt scaling prior to conformal prediction calibration~\cite{platt1999probabilistic}.

\section*{Acknowledgements}
We thank Mark Jayson Doma and Jhon Cedric Arcilla for their help in obtaining and understanding DialogflowCX model output. We kindly thank the reviewers for their time and their useful comments, without which this work would not have been possible in its current form.

\bibliography{bibliography}

\newpage
\clearpage

\appendix
\onecolumn
\section{Appendix: Implementation Details}
\label{sec:appendix_implementation_details}
We used \code{python v3.10.9} with packages \code{numpy} and \code{pandas} for data manipulation and basic calculations, \code{matplotlib} to generate illustrations, \code{mapie} for conformal prediction and reproduced these results in Julia and the package \code{conformalprediction.jl}. We used the \code{huggingface} API for fine tuning a version of \code{bert-base-uncased} using the hyperparameters below. For an anonymized version of the code and data see \url{https://anonymous.4open.science/r/cicc-205A}.

\begin{lstlisting}
learning_rate = 4.00e-05
warmup_proportion = 0.1
train_batch_size = 32
eval_batch_size = 32
num_train_epochs = 5
\end{lstlisting}

\subsection{Generative Language Model}
We use the \code{eachadea/vicuna-7b-1.1} variant of the LLAMA model using the HuggingFace API for the experiments presented here. We here provide an example prompt:
\begin{verbatim}
Customers asked an ambiguous question. Complete each set with a disambiguation question.

Set 1: Customer Asked: 'The terminal I paid at wouldn't take my card. Is something wrong?'
Option 1: 'card not working'
Option 2: 'card swallowed'
Disambiguation Question: 'I understand this was about you card. Was is swallowed or not working?'
**END**

Set 2:
Customer Asked: 'I have a problem with a transfer. It didn't work. Can you tell me why?'
Option 1: 'declined transfer'
Option 2: 'failed transfer'
Disambiguation Question: 'I see you are having issues with your transfer. Was your transfer failed or declined?'
**END**

Set 3: Customer Asked: 'I transferred some money but it is not here yet'
Option 1: 'balance not updated after bank transfer'
Option 2: 'transfer not received by recipient'
Disambiguation Question:
\end{verbatim}
\noindent More efforts can be spent on prompt engineering and more advanced generative LMs can be used, which we expect to improve the user satisfaction of CICC. Alternatively, simple text templates can be used. We consider the following alternatives and list some of their expected benefits and downsides:
\begin{description}
\item[Templates] a simple template-based can be used in which the user is asked to differentiate between the identified intents. Benefits of templates include full control over the chatbot output but a downside is that the CQs will be less varied, possibly sounding less natural and will not refer back to the users' original utterance,
\item[LM without user input] when using a LM, it is possible to not incorporate the user input $X$ in the prompt. This has the benefit of blocking any prompt injection but the downside of possibly unnatural CQs due to the inability to refer to the user query,
\item[LM with user input] by incorporating the user utterance into the LM prompt for CQ generation, the CQ can refer back to the user's phrasing and particular question, and therefore be formulated in a possibly more natural way.
\end{description}
We believe that more research is warranted to identify which of these approaches is most applicable in which cases, and how possible downsides of these alternatives can be mitigated in practice.

\section{Appendix: Sample ambiguous inputs}
\label{sec:appendix_ambiguous_inputs}
Tables~\ref{tab:ambiguous_inputs2}-~\ref{tab:ambiguous_inputs1} list inputs that are considered ambiguous by CICC in the B77 and HWU64 data sets respectively. Some inputs could refer to multiple intents whereas some other inputs could be considered out-of-scope.

\begin{table}[h!]
    \centering
    \footnotesize
    \begin{tabular}{l p{.2\textwidth} p{.125\textwidth} p{.575\textwidth}}

    \toprule
    \# & Utterance & Label & Prediction Set \\
    \midrule
1 & what is the matter? & direct debit payment not recognised & activate my card, age limit, balance not updated after bank transfer, balance not updated after cheque or cash deposit, beneficiary not allowed, cancel transfer, card arrival, card delivery estimate, card not working, card swallowed, cash withdrawal not recognised, change pin, compromised card, contactless not working, country support, declined card payment, declined transfer, direct debit payment not recognised, exchange rate, failed transfer, get physical card, lost or stolen card, lost or stolen phone, pending card payment, pending cash withdrawal, pending transfer, pin blocked, Refund not showing up, reverted card payment?, terminate account, top up failed, top up reverted, transaction charged twice, transfer not received by recipient, transfer timing, unable to verify identity, why verify identity, wrong amount of cash received,  \\
2 & Can I choose when my card is delivered? & card delivery estimate & activate my card, card about to expire, card acceptance, \underline{card arrival}, \underline{card delivery estimate}, change pin, contactless not working, country support, \underline{get physical card}, \underline{getting spare card}, getting virtual card, lost or stolen card, \underline{order physical card}, supported cards and currencies, top up by bank transfer charge, top up by card charge, visa or mastercard \\
3 & My contanctless has stopped working & contactless not working & activate my card, apple pay or google pay, automatic top up, beneficiary not allowed, cancel transfer, \underline{card not working}, card payment wrong exchange rate, \underline{contactless not working}, declined card payment, disposable card limits, failed transfer, get disposable virtual card, get physical card, pending top up, pin blocked, top up failed, top up reverted, topping up by card, virtual card not working, visa or mastercard, wrong exchange rate for cash withdrawal \\
4 & I misplaced my card and I dont know where the last place is where I used the card last. Can you look at my account and tell me the last place I used the card? & lost or stolen card & activate my card, atm support, card acceptance, card linking, card swallowed, cash withdrawal not recognised, \underline{compromised card}, \underline{lost or stolen card}, lost or stolen phone, order physical card, pin blocked \\
5 & Is my card denied anywhere? & card acceptance & \underline{atm support}, \underline{card acceptance}, card not working, card payment fee charged, card swallowed, compromised card, contactless not working, declined card payment, lost or stolen card, lost or stolen phone, order physical card, unable to verify identity, visa or mastercard \\
         \bottomrule
    \end{tabular}
    \caption{A sample of prediction sets on B77 of size $>th$ of seven with marginal conformal prediction on BERT outputs. Plausible labels have been highlighted with \underline{underscore}.}
    \label{tab:ambiguous_inputs2}
\end{table}

\begin{table}[h!]
    \centering
    \footnotesize
    \begin{tabular}{l p{.2\textwidth} p{.125\textwidth} p{.575\textwidth}}

    \toprule
    \# & Utterance & Label & Prediction Set \\
    \midrule
1 & olly & recommendation events & calendar set, general quirky, lists createoradd, music likeness, music query, play game, play music, play radio,  \\
2 & this song is too good & music likeness & audio volume mute, general affirm, general commandstop, general joke, general negate, lists remove, music dislikeness, \underline{music likeness} \\
3 & do i have to go to the gym & general quirky & \underline{calendar query}, \underline{general quirky}, lists query, \underline{recommendation events}, recommendation locations, transport traffic, weather query\\
4 & silently adjust & audio volume mute & \underline{audio volume down}, \underline{audio volume other}, \underline{audio volume up}, \underline{iot hue lightchange}, \underline{iot hue lightdim}, \underline{iot hue lightup}, music settings \\
5 & how many times does it go & general quirky & datetime query, \underline{general quirky}, \underline{lists query}, qa factoid, qa maths, \underline{transport query}, \underline{transport traffic} \\
6 & sports head lines please & news query & calendar set, general quirky, iot hue lightchange, music likeness, news query, qa factoid, social post, weather query \\
7 & read that back & play audiobook & email addcontact, email query, email querycontact, email sendemail, general quirky, lists createoradd, music likeness, play audiobook, play music, social post, \\
8 & i don't want to hear any more songs of that type & \underline{music dislikeness} & audio volume mute, calendar remove, general commandstop, iot wemo off, lists remove, music dislikeness, music likeness \\
9 & check celebrity wiki & general quirky & email query, \underline{general quirky}, \underline{lists query}, news query, \underline{qa factoid}, social post, social query \\
10 & Get all availables & lists query & email addcontact, email query, email querycontact, email sendemail, social post, social query, takeaway order,  \\
11 & rating & music likeness & cooking recipe, general quirky, lists createoradd, lists query, music likeness, music query, qa definition, qa factoid,  \\
12 & take me to mc donalds & transport query & play game, play podcasts, recommendation events, recommendation locations, recommendation movies, takeaway order, takeaway query \\
13 & search & qa factoid & email querycontact, general quirky, lists createoradd, lists query, music query, qa definition, qa factoid,  \\
14 & unmute & audio volume up & audio volume down, \underline{audio volume mute}, \underline{audio volume up}, iot wemo off, music settings, play radio, transport query, transport traffic \\
15 & please unmute yourself & audio volume mute & alarm remove, audio volume down, \underline{audio volume mute}, \underline{audio volume up}, iot cleaning, iot wemo on, music settings, play game \\
16 & what's the best day next week to go out for pizza & datetime query & \underline{calendar query}, cooking recipe, general quirky, qa factoid, \underline{recommendation events}, recommendation locations, \underline{takeaway query} \\
17 & i need a manger & general quirky & calendar set, cooking recipe, general quirky, lists createoradd, music likeness, play game, qa definition, qa factoid, social post,  \\
18 & assistant shuffle entire library & play music & iot cleaning, iot hue lightchange, lists createoradd, \underline{music settings}, \underline{play audiobook}, play game, \underline{play music} \\
19 & put the disco lights on & iot hue lighton & alarm remove, iot cleaning, \underline{iot hue lightchange}, iot hue lightoff, \underline{iot hue lighton}, \underline{iot hue lightup}, iot wemo on \\
20 & hello how are you today & general greet & \underline{general greet}, general praise, \underline{general quirky}, play radio, recommendation events, recommendation locations, recommendation movies \\
21 & where does tar work currently & email querycontact & cooking recipe, \underline{email querycontact}, general quirky, lists query, qa definition, recommendation locations, takeaway query \\
22 & can you pull up jeff & email querycontact & general quirky, iot cleaning, news query, play game, play music, play radio, social post \\
23 & show me theatre where pirates of caribbean is playing & recommendation movies & general quirky, play game, play music, play podcasts, recommendation events, recommendation locations, \underline{recommendation movies} \\
24 & spot & play podcasts & calendar set, music likeness, music query, play music, play radio, qa factoid, weather query,  \\
25 & anything i need to know & general quirky & \underline{calendar query}, calendar set, \underline{email query}, email querycontact, \underline{general quirky}, lists query, \underline{news query}, \underline{social query} \\
26 & give me some good sounds & play music & audio volume down, \underline{audio volume up}, music likeness, \underline{music query}, music settings, \underline{play music}, \underline{play radio} \\
27 & i don't like it & music dislikeness & calendar remove, general commandstop, \underline{general dontcare}, general joke, \underline{general negate}, lists remove, \underline{music dislikeness}, music likeness, social post \\
28 & what are you & general quirky & audio volume down, audio volume mute, audio volume up, music query, music settings, qa definition, recommendation locations, recommendation movies \\
29 & i don't like this one & general quirky & \underline{audio volume mute}, calendar remove, general joke, \underline{general negate}, \underline{iot wemo off}, \underline{lists remove}, \underline{music dislikeness}, social post \\
         \bottomrule
    \end{tabular}
    \caption{Prediction sets on ACID of size $>th$ of seven with marginal conformal prediction on BERT outputs. Plausible labels have been highlighted with \underline{underscore}.}
    \label{tab:ambiguous_inputs1}
\end{table}

\section{Appendix: LLM results}
We here present a random sample of CQs on B77 and C150.

\begin{table}[h!]
 \centering
    \begin{tabular}{p{.3\textwidth} p{.2\textwidth} p{.5\textwidth}}
    \toprule
    Utterance & Prediction Set & LLM output \\
    \midrule
I hope you can help me. My account has been compromised from a different town that I haven't been to. Can you please place a hold on my account until I can make it to the bank? & cash withdrawal not recognised, compromised card & I understand you are having issues with your account. Was your account compromised or your card not recognised for cash withdrawal?\\
Is my card denied anywhere? & card not working, declined card payment & I understand you are asking about your card. Was it declined or not working?\\
Money that I have transferred hasn't arrived yet & balance not updated after bank transfer, transfer not received by recipient & I understand you are waiting for a transfer to be updated in your balance. Was the transfer not received by you or the recipient?\\
How long will the transfer take? & balance not updated after bank transfer,pending transfer, transfer timing & I understand you are asking about transfer timing. Was your transfer pending, not updated or not received?\\
What should I do with my stuck ATM? & card swallowed, declined cash withdrawal & I understand you have a stuck ATM. Was it your card that got stuck or was your cash withdrawal declined?\\
\bottomrule
\end{tabular}
\caption{Sample of LLM outputs on the B77 dataset where the user input has been made part of the prompt.}
\label{}
\end{table}
\clearpage

\begin{table}[h!]
 \centering
    \begin{tabular}{p{.3\textwidth} p{.2\textwidth} p{.5\textwidth}}
    \toprule
    Utterance & Prediction Set & LLM output \\
    \midrule
how many more miles can i drive on my current tank & gas, mpg & I understand you are asking about your car's fuel. Are you asking about the number of miles you can drive on your current tank or the miles per gallon?\\
can i redeem my points for flights & redeem rewards, rewards balance & I understand you would like to redeem your points for flights. Are you looking to check your rewards balance or redeem your rewards?\\
are there any meetings between 2 and 5 pm & schedule meeting, meeting schedule & I understand you are looking for a schedule of meetings. Are you looking for a specific time range or a general schedule?\\
is the marathon on friday added to my calendar & calendar, calendar update & I understand you are referring to your calendar. Is it the marathon on friday that you are referring to or do you need an update on your calendar?\\
why didn't my card work & expiration date, card declined & I understand you are having issues with your card. Is it because of the expiration date or was it declined?\\
\bottomrule
\end{tabular}
\caption{Sample of LLM outputs on the C150 dataset.}
\end{table}

\clearpage

\section{Appendix: Intent distributions}
We here present the intent distributions for all public datasets across train, calibration and test splits.

\begin{figure}[h!]
\centering
\includegraphics[width=.7\textwidth]{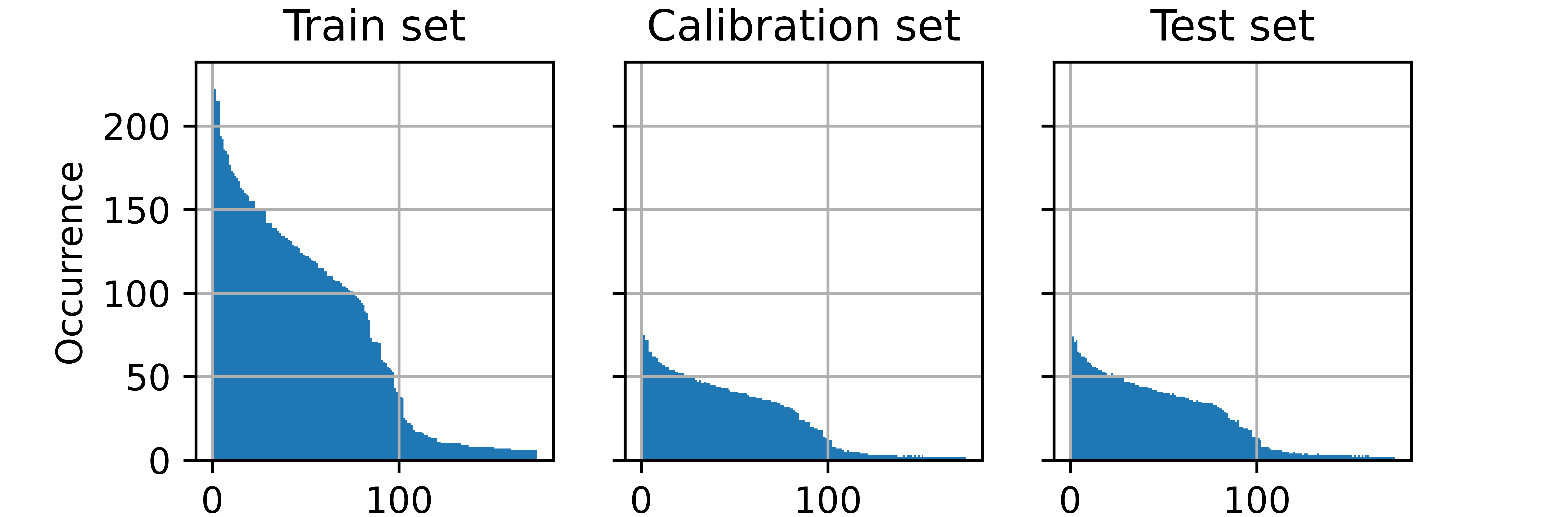}
\caption{Intent distribution in ACID data set.}
\end{figure}

\begin{figure}[h!]
\centering
\includegraphics[width=.7\textwidth]{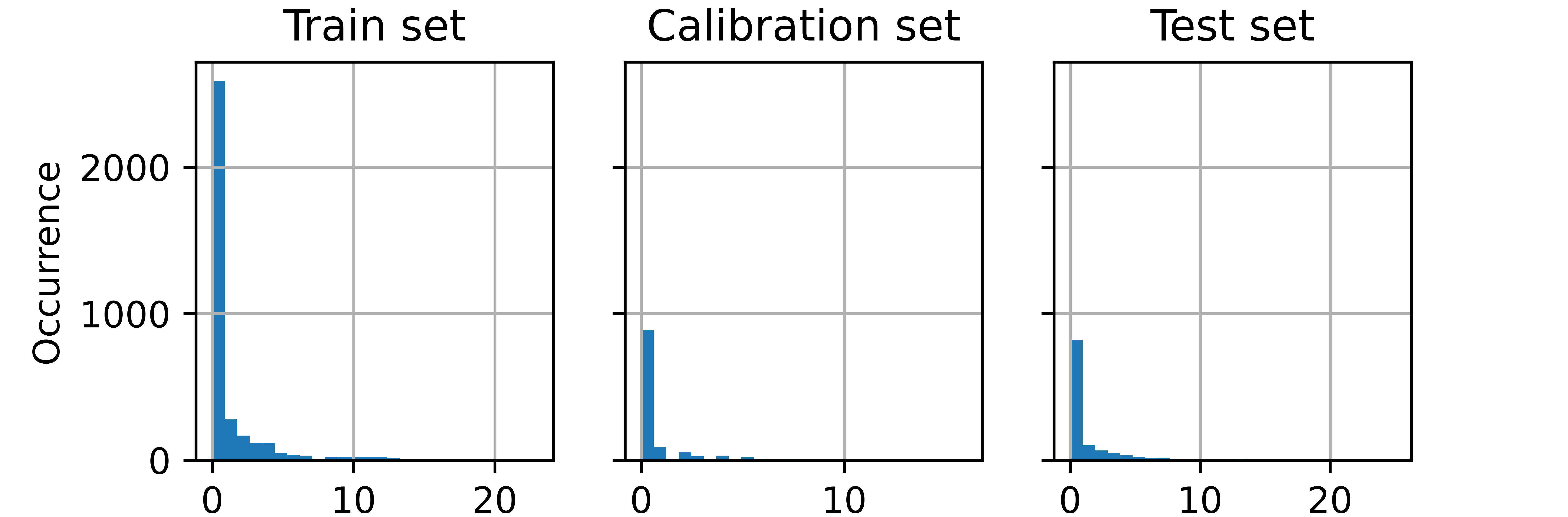}
\caption{Intent distribution in ATIS data set.}
\end{figure}

\begin{figure}[h!]
\centering
\includegraphics[width=.7\textwidth]{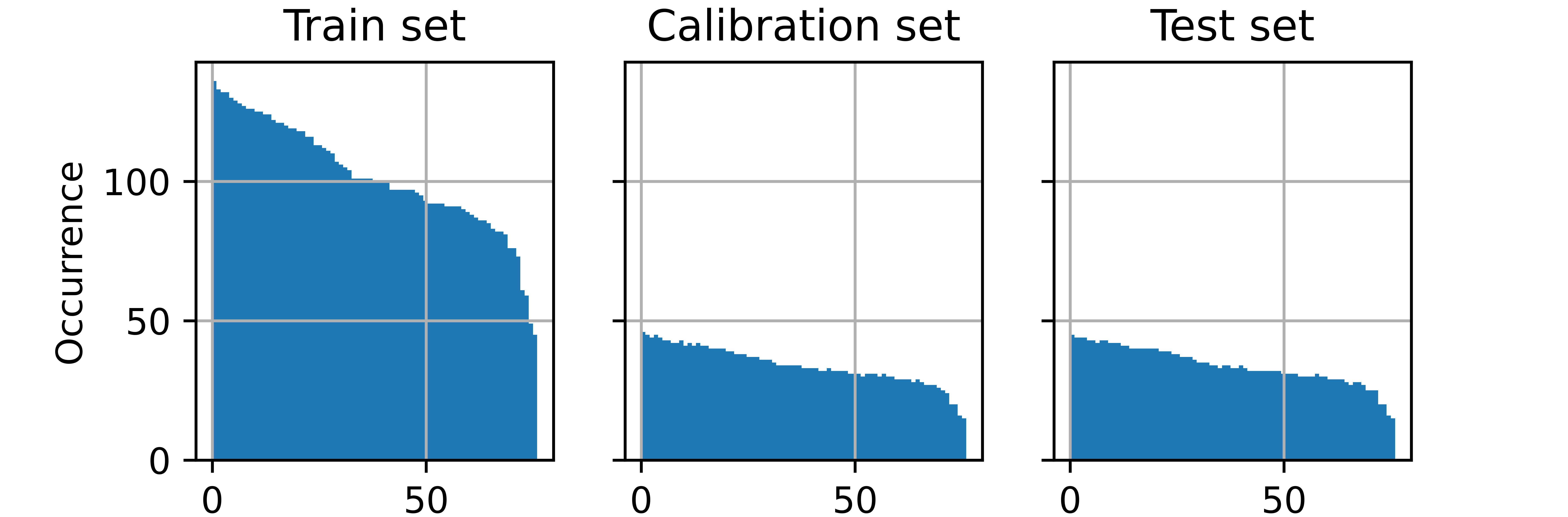}
\caption{Intent distribution in B77 data set.}
\end{figure}

\begin{figure}[h!]
\centering
\includegraphics[width=.7\textwidth]{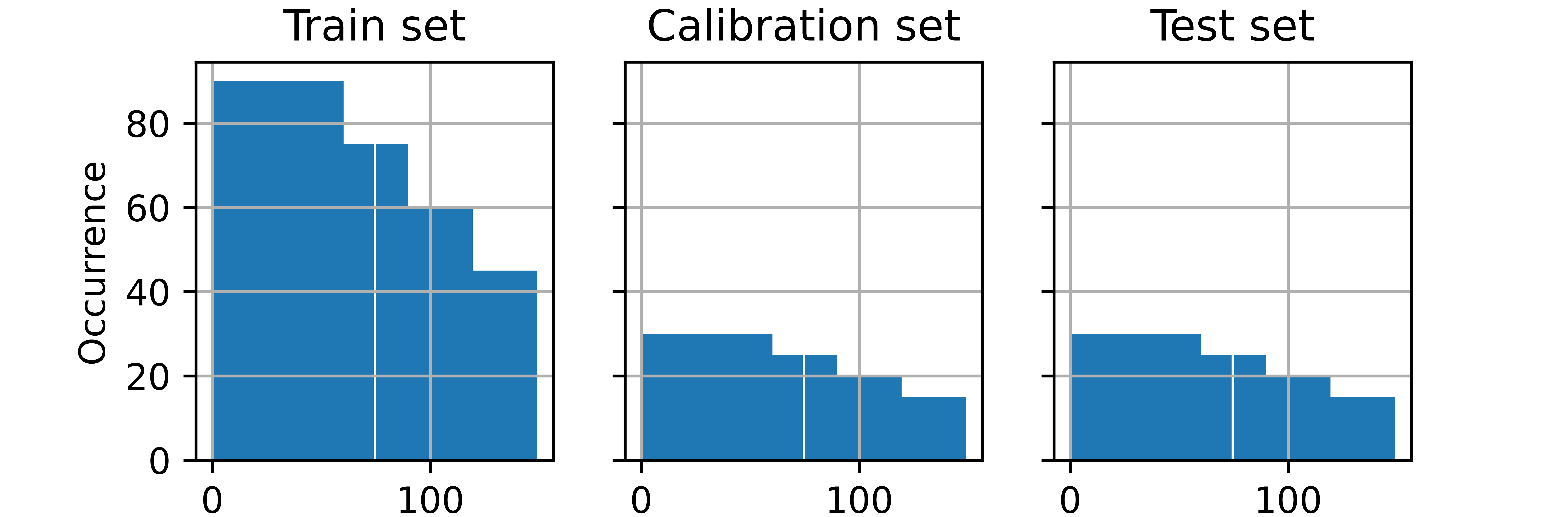}
\caption{Intent distribution in C150-IS data set.}
\end{figure}

\begin{figure}[h!]
\centering
\includegraphics[width=.7\textwidth]{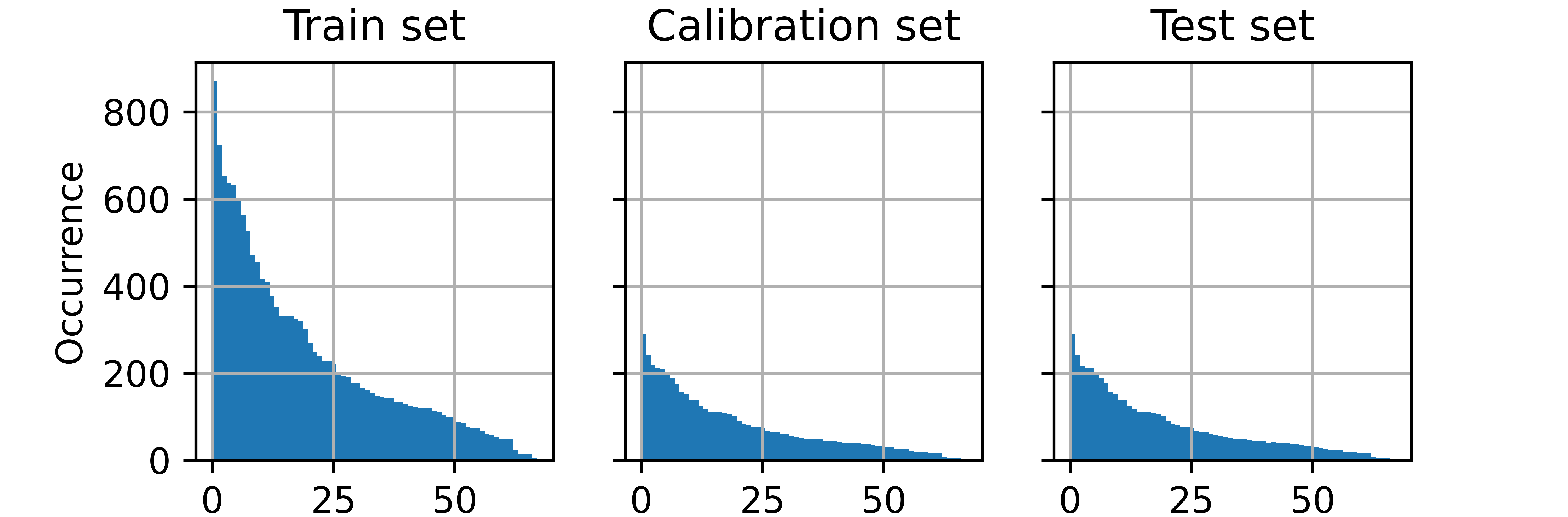}
\caption{Intent distribution in HWU64 data set.}
\end{figure}

\begin{figure}[h!]
\centering
\includegraphics[width=.7\textwidth]{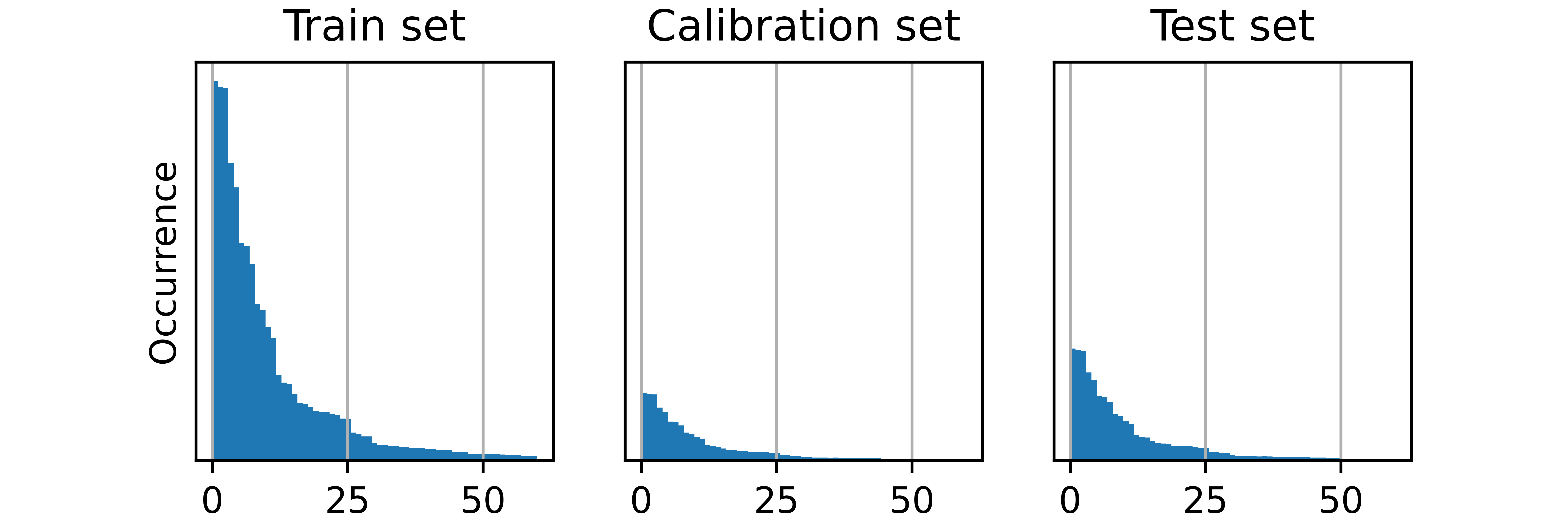}
\caption{Intent distribution in IND data set.}
\end{figure}

\begin{figure}[h!]
\centering
\includegraphics[width=.7\textwidth]{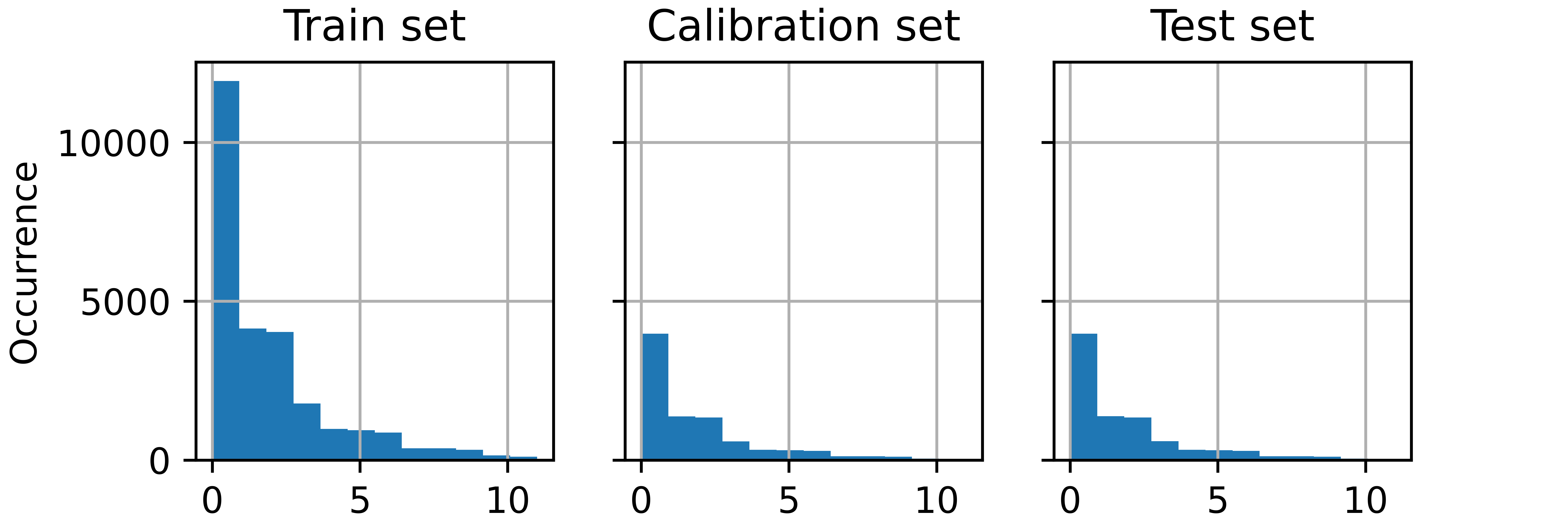}
\caption{Intent distribution in MTOD data set.}
\end{figure}

\clearpage

\section{Appendix: Unoptimized $\alpha$}
This appendix contains results for an unoptimized $\alpha$ hyperparameter, arbitrarily set at $.10$ and $.01$. We see that for most data sets, there is no need to ask a clarification question as the model already achieves the desired coverage. Much higher coverages (as in Table~\ref{tab:results}) are achievable for these data sets. For some more challenging data sets such as C150, HWU64 and IND, CICC yields small clarification questions while retaining a reasonably large number of clarification questions of size 1.
\begin{table}[h!]
\centering
\footnotesize{
\setlength\tabcolsep{2pt}
\begin{tabular}{lccr|cccc}
\toprule
Setting & $1-\alpha$ & $th$ &    & Cov$\uparrow$ & Single$\uparrow$ & $|\text{CQ}|\downarrow$ & Amb \\ \midrule
ACID & .90 & 7  &  CICC & \underline{.90} & .92 & $-$ &0 \\
& & &  B1 & \underline{.97} & .93 & 5 &0 \\
& & &  B2 & \underline{.95} & \textbf{1} & $-$ &0 \\
& & &  B3 & \underline{.99} &0 & 5 &0 \\
\midrule
ATIS & .90 & 7  &  CICC & .88 & .89 & $-$ &0 \\
& & &  B1 & \underline{.99} & .93 & 5 &0 \\
& & &  B2 & \underline{.98} & \textbf{1} & $-$ &0 \\
& & &  B3 & \underline{1} &0 & 5 &0 \\
\midrule
B77/BERT & .90 & 7  &  CICC & \underline{.98} & .79 & \textbf{2.90} & .04 \\
& & &  B1 & \underline{.97} & .90 & 5 &0 \\
& & &  B2 & \underline{.93} & \textbf{1} & $-$ &0 \\
& & &  B3 & \underline{.99} &0 & 5 &0 \\
\midrule
B77/DFCX & .90 & 4  &  CICC & \underline{.91} & .66 & 2.63 & .02 \\
& & &  B1 & \underline{.95} & .71 & 4.79 & .27 \\
& & &  B2 & .90 & \textbf{.98} & \textbf{2.26} &0 \\
& & &  B3 & \underline{.97} &0 & 5 & 1 \\
\midrule
C150 & .90 & 7  &  CICC & \underline{.99} & .97 & \textbf{2.66} &0 \\
& & &  B1 & \underline{.99} & .82 & 5 &0 \\
& & &  B2 & \underline{.98} & \textbf{1} & $-$ &0 \\
& & &  B3 & \underline{1} &0 & 5 &0 \\
\midrule
HWU64 & .90 & 7  &  CICC & \underline{.90} & .97 & \textbf{2.00} &0 \\
& & &  B1 & \underline{.96} & .79 & 5 &0 \\
& & &  B2 & \underline{.90} & \textbf{1} & $-$ &0 \\
& & &  B3 & \underline{.98} &0 & 5 &0 \\
\midrule
IND & .90 & 7  &  CICC & \underline{.91} & \textbf{.25} & \textbf{3.46} & .11 \\
& & &  B1 & .88 & .42 & 5 &0 \\
& & &  B2 & .70 & 1 & $-$ &0 \\
& & &  B3 & \underline{.91} &0 & 5 &0 \\
\midrule
MTOD & .90 & 7  &  CICC & \underline{.90} & .90 & $-$ &0 \\
& & &  B1 & \underline{.99} & .99 & 5 &0 \\
& & &  B2 & \underline{.99} & \textbf{1} & $-$ &0 \\
& & &  B3 & \underline{1} &0 & 5 &0 \\
 
\bottomrule
\end{tabular}}
    \caption{Test set results for $1-\alpha=.90$ where \underline{underline} indicates meeting coverage requirement. \textbf{Bold} denotes best when meeting this requirement, omitted for last column due to missing ground truth for ambiguous.}
    \label{tab:results_unoptimized}
\end{table} 
\begin{table}[h!]
\centering
\footnotesize{
\setlength\tabcolsep{2pt}
\begin{tabular}{lccr|cccc}
\toprule
Setting & $1-\alpha$ & $th$ &    & Cov$\uparrow$ & Single$\uparrow$ & $|\text{CQ}|\downarrow$ & Amb \\ \midrule
\midrule
ACID & .99 & 7   &  CICC & \underline{1} & \textbf{.77} & \textbf{3.00} & .10 \\
& & &  B1 & .98 & .85 & 5 & 0 \\
& & &  B2 & .95 & 1 & $-$ & 0 \\
& & &  B3 & \underline{.99} & 0 & 5 & 0 \\
\midrule
ATIS & .99 & 7  &  CICC & \underline{.99} & \textbf{.98} & \textbf{2.54} & 0 \\
& & &  B1 & \underline{.99} & .73 & 5 & 0 \\
& & &  B2 & .98 & 1 & $-$ & 0 \\
& & &  B3 & \underline{1} & 0 & 5 & 0 \\
\midrule
B77/BERT & .99 & 7  &  CICC & \underline{.98} & \textbf{.79} & \textbf{2.90} & .04 \\
& & &  B1 & .97& .90 & 5 & 0 \\
& & &  B2 & .93 & 1 & $-$ & 0 \\
& & &  B3 & \underline{.99} & 0 & 5 & 0 \\
\midrule
B77/DFCX & .99 & 4  &  CICC & .97 & 0 & 5 & 1 \\
& & &  B1 & .97 & .05 & 5 & .95 \\
& & &  B2 & .90 & 1 & $-$ & 0 \\
& & &  B3 & .97 & 0 & 5 & 1 \\
\midrule
C150 & .99 & 7  &  CICC & \underline{.99} & \textbf{.97} & \textbf{2.66} & 0 \\
& & &  B1 & \underline{.99} & .82 & 5 & 0 \\
& & &  B2 & .98 & 1 & $-$ & 0 \\
& & &  B3 & \underline{1} & 0 & 5 & 0 \\
\midrule
HWU64 & .99 & 7  &  CICC & \underline{.99} & \textbf{.25} & \textbf{3.39} & .28 \\
& & &  B1 & .98 & .05 & 5 & 0 \\
& & &  B2 & .90 & 1 & $-$ & 0 \\
& & &  B3 & .98 & 0 & 5 & 0 \\
\midrule
MTOD & .99 & 7  &  CICC & \underline{.99} & \textbf{1} & $-$ & 0 \\
& & &  B1 & \underline{1} & .98 & 5 & 0 \\
& & &  B2 & \underline{.99} & \textbf{1} & $-$ & 0 \\
& & &  B3 & \underline{1} & 0 & 5 & 0 \\
\bottomrule
\end{tabular}}
    \caption{Test set results for $1-\alpha=.99$ where \underline{underline} indicates meeting coverage requirement. \textbf{Bold} denotes best when meeting this requirement, omitted for last column due to missing ground truth for ambiguous.}
    \label{tab:results_unoptimized2}
\end{table}

\clearpage

\section{Appendix: Comparison results OOS detection}
We here compare the results of OOS detection as reported by baselines. Note that these results were generated on different splits of the data and (where applicable), possibly using different open-domain samples, and that a direct comparison between results is invalid.
\begin{table}[h!]
\centering
% \footnotesize{
\setlength\tabcolsep{2pt}
\begin{tabular}{ll|cc}
\toprule
Dataset         &  Algorithm                & F1$\uparrow$    & Accuracy$\uparrow$ \\
\midrule
C150            &   CICC-OOS                    & .91   & .68 \\ 
                &   \citet{zhan2021out} 25\%   & .81   & .88 \\
                &   \citet{zhan2021out} 50\%   & .87   & .88 \\
                &   \citet{zhan2021out} 75\%   & .89   & .88 \\
                &   \citet{cavalin2020improving}& .76  & .73 \\ \midrule
B77             &   CICC-OOS                    & .90   & .89 \\
                &   \citet{zhan2021out} 25\%   & .74   & .70 \\
                &   \citet{zhan2021out} 50\%   & .80   & .73 \\
                &   \citet{zhan2021out} 75\%   & .87   & .81 \\
\bottomrule
\end{tabular}
    \caption{Results for the OOS detection task.}
    \label{tab:ood_results_app}
\end{table}

\end{document}